# Fast TILs - A Pipeline for Efficient TILs Estimation in Non-Small Cell Lung Cancer

Nikita Shvetsov[1]*, Anders Sildnes[1], Masoud Tafavvoghi[2], Lill-Tove Rasmussen Busund[3,4], Stig Manfred Dalen[4], Kajsa Møllersen[2], Lars Ailo Bongo[1], Thomas K. Kilvaer[5,6]
1 Department of Computer Science, UiT The Arctic University of Norway.
2 Department of Community Medicine, UiT The Arctic University of Norway.
3 Department of Medical Biology, UiT The Arctic University of Norway.
4 Department of Clinical Pathology, University Hospital of North Norway.
5 Department of Oncology, University Hospital of North Norway.
6 Department of Clinical Medicine, UiT The Arctic University of Norway.
* Corresponding author: nikita.shvetsov@uit.no

## Abstract

**Background**: The prognostic relevance of tumor-infiltrating lymphocytes (TILs) in non-small cell lung cancer (NSCLC) is well-established. However, manual TIL quantification in hematoxylin and eosin (H&E) whole slide images (WSIs) is laborious and prone to variability. To address this, we aim to develop and validate an automated computational pipeline for the quantification of TILs in WSIs of NSCLC. Such a solution in computational pathology can accelerate TIL evaluation, thereby standardizing the prognostication process and facilitating personalized treatment strategies.

**Methods**: We developed an end-to-end automated pipeline for TIL estimation in lung cancer WSIs, by integrating a patch extraction approach based on hematoxylin component filtering with a machine learning-based patch classification and cell quantification method using the HoVer-Net model architecture. Additionally, we employ randomized patch sampling to further reduce the processed patch amount. We evaluated the patch sampling procedure effectiveness, the pipelines' ability to identify informative patches and computational efficiency, and clinical value of produced scores using patient survival data.

**Results**: Our pipeline demonstrates the ability to selectively process informative patches, achieving a balance between computational efficiency and prognostic integrity. The pipeline filtering excluded approximately 70% of all patch candidates. Further, only 5% of eligible patches were necessary to retain the pipeline's prognostic accuracy (c-index = 0.65) resulting in a linear reduction of the total computational time compared to the filtered patch subset analysis. The pipeline's TILs score had a strong association with patient survival, and outperformed traditional CD8 immunohistochemical scoring (c-index = 0.59). Kaplan-Meier analysis further substantiated the TILs score's prognostic value.

**Conclusion**: This study introduces an automated pipeline for TIL evaluation in lung cancer WSIs, providing a prognostic tool with potential to improve personalized treatment in NSCLC. The pipeline's computational advances, particularly in reducing processing time, and clinical relevance demonstrate a step forward in computational pathology.

# 1. Introduction

Computational pathology is an emerging field set to revolutionize cancer detection and prognostication. Alongside DNA, RNA, radiomics and proteomics, computational pathology is making significant contributions to personalized cancer treatment and has the potential for novel biomarker discovery in clinical settings by shifting from traditional, manual microscope analyses to advanced digital techniques that leverage machine learning and big data [1], [2].

Tumor-infiltrating lymphocytes (TILs) are significant prognostic and potentially predictive biomarkers in numerous cancers [3], [4], [5]. Unfortunately, manual TIL quantification in whole slide images (WSIs) is labor-intensive, subject to personal bias, and prone to inaccuracies. To reduce the workload of pathologists, a common approach is to estimate the impact of TILs by their total number on a coarse scale or by subjective morphological descriptions [6], [7]. However, even these simple approaches can be time-consuming and may not accurately represent the true condition of the disease. Hence, TIL quantification is one of the processes where computational pathology may impact clinical implementation and lead to improved prognostication for cancer patients.

Although possible, performing detailed computations across all regions of a WSI is impractical due to 1) the immense data volume leading to extended computation times, 2) potential degradation of the assessment's performance due to the human factor introduced by selective "cherry-picking" from all processed patches, and 3) the extensive hardware and electricity demands associated with processing such large datasets.

To address these challenges, we present an automated pipeline for computationally efficient TIL evaluation in WSIs. Recognizing that not all areas within a WSI hold equal prognostic value, our pipeline employs a randomized patch sampling strategy. This approach selectively targets areas with high cell content, samples a fraction of those areas, and filters out the less relevant patches to retain only those with prognostic relevance, thereby avoiding the exhaustive analysis of entire slides. By focusing our computational efforts on selected patches, we significantly reduce the time and computational costs, which is crucial for practical applications in clinical and research settings. Our method ensures that the analysis remains reproducible and scalable, while maintaining the integrity and accuracy of the prognostic evaluation.

This work builds on our previous research, which introduced a pragmatic machine learning methodology for quantifying TILs in tissue from non-small cell lung cancer (NSCLC) patients [8]. Our initial results suggested that our method's prognostic impact could be comparable or even superior to the current standard CD8 immunohistochemical (IHC) staining methods for TIL evaluation in NSCLC. However, in this work we relied on manual selection of relevant patches to evaluate our model, thereby negating many of the potential benefits of computational pathology and introducing a potential selection bias. Herein, we refine this approach by developing an end-to-end pipeline including an automated patch extraction and patch classification model for identifying prognostically relevant patches.

Our contribution is an efficient and robust automated pipeline for evaluating TILs in WSIs from NSCLC patients. The pipeline presented here aims to enhance the accuracy and consistency of TILs scoring by mitigating potential human error and bias. Furthermore, it generates detailed visualizations that highlight areas with high TILs density, offering pathologists insightful information to pinpoint areas of interest and make more informed decisions. In conclusion, our automated end-to-end method for TILs evaluation in WSIs can improve personalized NSCLC treatment.

# 2. Methods

## 2.1 Datasets

In this study, we employed three datasets that we created or modified for the development and evaluation of our pipeline: the UNN-NSCLC WSI dataset, the UNN-LC patch dataset, and the TCGA validation dataset. The WSI data distribution and relationships among these datasets are shown in Appendix Figure A1.1. We also use the PanNuke dataset [9] for cell quantification model training and testing.

### 2.1.1 UNN-NSCLC WSI dataset

The primary dataset comprises clinical information and tissue from 553 NSCLC patients treated at the University Hospital of North Norway and Nordland Central Hospital from 1990 to 2010. This cohort has previously been extensively described [10], and information on different immune cell subsets is available [11]. Of these 553 patients, we use the 497 who have an available H&E WSI, a complete set of clinical variables, and a CD8 score. The WSIs were captured using an Aperio scanner at x40 magnification and are approximately 10^6 x 10^6 pixels at highest resolution. Moreover, these WSIs come with slide-level annotations of viable tumor areas, performed by an experienced pathologist as part of the doMore! project [12]. The annotations, stored as polygons in .xml files, outline the viable tumor areas within each WSI. While the annotations were not directly used in this project, they served as a valuable reference for understanding the overall tumor landscape within the WSIs.

It is important to note that we do not possess a 'ground truth' in terms of labeled cells or a reference TIL density per patch or WSI. Ascertaining a ground truth would require the labeling of millions of cell instances across the 497 WSIs, which is not practically feasible. Therefore, clinical data serves as our reference point for comparison and validation without a precise ground truth.

### 2.1.2 UNN-LC patch dataset

Our secondary dataset is an assembly of randomly selected patches from a subset of WSIs derived from the UNN-NSCLC WSI dataset and lung cancer patches from the LC25000 dataset [13]. Each patch is 768 x 768 pixels. The labeled part of dataset was created by an oncologist (TKK) in collaboration with an experienced pathologist (SD) using QuPath [14] and a custom-developed patch annotation tool [15], specifically designed for this project. The process of patch selection and labeling was designed to mitigate potential selection and labeling bias. The labeled data generated from UNN-NSCLC WSI data comprises 1628 necrosis, 1913 stroma, 1962 normal lung, and 1069 tumor tissue patches.

To expand the labeled dataset, additional patches were incorporated from the LC25000 dataset. This expansion included 5000 patches each of lung adenocarcinoma, lung squamous cell carcinoma, and benign lung tissue. We integrated the lung adenocarcinoma and lung squamous cell carcinoma patches into the tumor tissue class, and the benign lung tissue patches into the normal lung class, ensuring a robust and varied dataset for analysis.

### 2.1.3 TCGA validation dataset

To evaluate the generalizability and robustness of our model, we conduct an external validation using The Cancer Genome Atlas (TCGA) datasets: TCGA-LUAD (lung adenocarcinoma) and TCGA-LUSC (lung squamous cell carcinoma). For clinical and meta-information, we use an updated list of clinical variables related to patient survival [16]. The initial set of WSIs was filtered to include only slides that contained a disease label. Additionally, we only include patients that had a follow-up time of at least 30 days, whose tumor was not classified as stage IV and had an available diagnostic slide without prevalent necrotic areas. Upon assessing the scanning resolutions in the datasets, we exclude the 20x resolution images due to insufficient quality. Consequently, the final dataset comprise 387 WSIs for TCGA-LUAD and 382 WSIs for TCGA-LUSC, all of which were scanned at 40x resolution.

## 2.2 Algorithms and models

To achieve the goal of automatic evaluation of TILs abundance in NSCLC WSIs, we have developed a pipeline, combining three consecutive steps: patch extraction, patch classification, and cell quantification (Figure 1).

These steps are executed fully automated, yielding visual and quantitative measures of TILs in WSIs. This information can aid clinicians in evaluating patients' prognoses, thereby enhancing the overall efficacy and precision of NSCLC diagnoses and treatment planning.

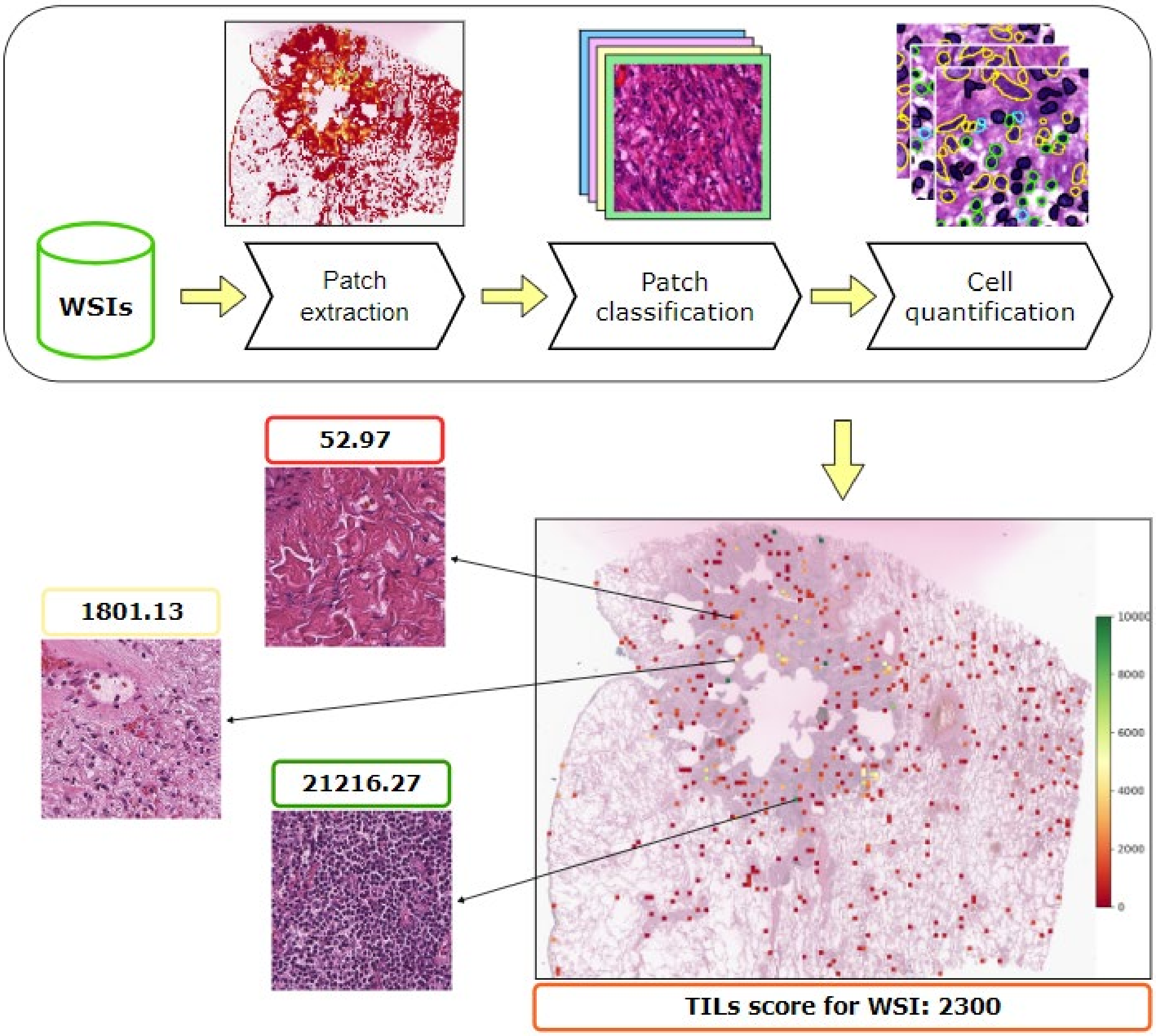

*Figure 1. Proposed pipeline for evaluating TILs density in WSI. Results include the heatmap visualization of processed patches and TILs score for WSI.*

### 2.2.1 Patch extraction for selecting regions with prognostic information

The aim of the first pipeline stage is to remove areas without tissue and cell content. The patch extraction step consists of two consecutive steps that filter out non-tissue area and patches without cell-content information. By doing so, we optimize the analysis to focus on clinically relevant regions, avoiding exhaustive computations further down the pipeline.

**Tissue extraction:** To extract the tissue area, we downsample the original WSI, convert the image to grayscale, and apply a binary threshold set at 170 to distinguish and eliminate non-tissue areas. We chose the threshold empirically by performing quality analysis on our UNN-NSCLC WSI dataset. Then we discard small tissue outliers and artifacts. We identify the remaining non-empty regions as our initial set of patch candidates. We calculate contours for those regions on the downsampled image and scale them back to the original resolution using an affine transformation. This technique rapidly isolates tissue areas without the need for a comprehensive analysis of the entire slide. After processing the contours, we compute a patch coordinate grid for the next sub-step to crop and evaluate areas patch by patch.

**H-filtering:** When a WSI is stained with hematoxylin and eosin (H&E), the hematoxylin stains the cell nuclei, while eosin stains other cytoplasmic components and the extracellular matrix. The intensity of the hematoxylin staining correlates with the number of cell nuclei in each area, which in turn reflects the cellular content of that tissue patch. By performing color deconvolution and transforming the RGB image to the Hematoxylin-Eosin-DAB (HED) color space [17], we assess the hematoxylin component within each patch. We then compare the mean hematoxylin channel value against a predetermined threshold of 0.017. The threshold is an empirical value derived by visually inspecting a random subset of patches from UNN-NSCLC WSI dataset and evaluating the hematoxylin component maps to ensure that we focus on regions with sufficient cellular content. It is important to note that this threshold may require adjustment when applying the pipeline to other datasets with different staining characteristics.

The output of this process is a table of coordinate pairs representing the selected patches, which optimizes both computational time and disk space usage. Adding the H-filtering approach on top of Tissue extraction significantly reduces the patch counts for the pipeline, contributing to the high-throughput capabilities of the pipeline.

### 2.2.2 Patch classification for identifying patches with prognostic information

The second step in the pipeline is patch classification. In NSCLC diagnostics, pathologists routinely ignore necrotic regions and normal lung tissue due to the minimal prognostic information they offer. These patches contain cells and are therefore not removed during the initial filtering in 2.2.1.

We develop a classification model comprising an EfficientNetV2 [18] backbone architecture (V2-S configuration) and a classification head with a dropout layer. The architecture's advanced use of Fused-MBConv blocks and a progressive learning strategy enables it to efficiently handle complex patterns and to emphasize prominent features within the detailed tissue images. Compared to previous iterations of EfficientNet family models, EfficientNetV2's improved scaling method and training speed optimizations align with the demands of histopathology, scaling up the network to process large images, ensuring detailed pattern recognition in tissue structures, while being parameter efficient and less computationally demanding during training.

For training and testing the patch classifier model, we use the UNN-LC patch dataset. We split the UNN-LC patch dataset into a training set comprising patches from 74 patients (patches derived from 68 patients of the UNN-NSCLC WSI dataset and 6 additional patients) and a test set comprising patches from 120 patients (patches derived from 113 patients from the UNN-NSCLC WSI dataset and 7 additional patients). The resulting training set consists of 1350 patches per class, with the remaining patches allocated to the testing set. Downloadable data and description of the UNN-LC patch dataset is available in the data repository [19].

The EfficientNetV2-based patch classifier model was trained using 5-fold cross-validation, with the same class balance in each fold. To mitigate the risk of overfitting, a series of augmentation techniques are employed. These techniques included affine transformations - translation, scaling, and rotation - to simulate variability in tissue sample positioning. Moreover, we randomly modify contrast, saturation, and brightness values of the patches to account for variations in histological processing. Additionally, we apply random normalization based on the Macenko algorithm [20], [21] to standardize the staining appearance across different samples.

During inference, we employ the trained model to process the list of patches. The output is a modified table from the patch extraction step 2.2.1, with a class label assigned to each patch. For the subsequent analysis we select only the patches that correspond to the 'tumor tissue' and 'stroma' classes, which include tumor tissue, stroma, tertiary lymphoid structures, and areas with TILs.

### 2.2.3 Cell quantification and TILs scoring

The third step in the pipeline is cell quantification. For each selected patch from the previous step, we extract its coordinates, retrieve the patch from the corresponding WSI, normalize patch pixel values and perform inference using a modified version of HoVer-Net [22] based model trained on the PanNuke dataset [9]. The choice of the PanNuke dataset was based on its diversity and the substantial amount of labeled cell nuclei. The trained model delineates cellular structures, classifies them, and quantifies TILs. This quantification pipeline has been adapted from our earlier study [8].

Upon quantifying the cells, we obtain raw TIL counts from within each analyzed patch (Figure 1). To enable comparison across different patches and WSIs, we normalize these counts by the area of the patch, accounting for the resolution of the WSI. The normalization formula (1), adjusted for square micrometers and standardized to cells per square millimeter, is as follows:

$$d_i = \frac{c_i}{(patch_size \times mpp)^2} \times 10^6 \quad (1)$$

Here, $d_i$ represents the density of TILs in cells per square millimeter for patch $i$, $c_i$ is the absolute TIL count for patch $i$, and $patch_size$ represents the uniform dimension of the patch in pixels, given that the patch is square with equal width and height. The term $mpp$ refers to the microns-per-pixel ratio of the WSI. The factor $10^6$ is applied to convert the area from square microns to square millimeters.

For a given patient, the final TIL density score, defined in equation (2), is computed as the mean density across all patches within the WSI. In our current cohort, each patient is represented by a single WSI, hence, the calculation of $d_{patient}$ does not require averaging across multiple slides:

$$d_{patient} = \frac{1}{n}\sum_{i=1}^{n} d_i \quad (2)$$

It is important to note that for studies with multiple WSIs per patient, the final TIL density score would be the average density across all patches and all $k$ WSIs for the patient, as represented by the formula (3):

$$d_{patient} = \frac{1}{k}\sum_{j=1}^{k} \left(\frac{1}{n_j}\sum_{i=1}^{n_j} d_i\right) \quad (3)$$

The output of this step is an aggregated TIL density scores for the WSI and per patch scores.

## 2.3 Randomized patch sampling

Although the pipeline quantifies lymphocytes in a fraction of the overall patch count of WSI, the patch extraction step (Section 2.2.1) still yields a substantial number of patches. We further explore optimization of this number for faster computational processing with informative results as an output.

To achieve this, we adopt a strategy of randomized sampling from the patch set, obtained after the patch extraction step, which enables a reduction in the volume of data passed to the classification model (Section 2.2.2) and subsequently to the quantification model (Section 2.2.3).

To parameterize and evaluate the patch sampling procedure we conduct empirical analysis on a subset of 50 WSIs, wherein the number of patches was systematically reduced using Monte-Carlo simulations after the patch extraction step and passed further down the pipeline. The iterative process aimed to identify a sampling ratio that maintains a prognostic value while optimizing computational efficiency.

## 2.4 Pipeline results visualization

The TIL scores for a single WSI and its processed patches can be visualized by assigning a TILs score for each patch and constructing a heatmap as visualized in Figure 2. In this WSI, the TILs score varies from 0 (no TILs) to ≥10000 (extremely high TILs density, clipped at 10000 to due to outliers). This visualization provides a comprehensive overview of the TILs distribution across the WSI, further enhancing pathologists' vision and understanding of the tumor microenvironment and its potential implications on patient prognosis.

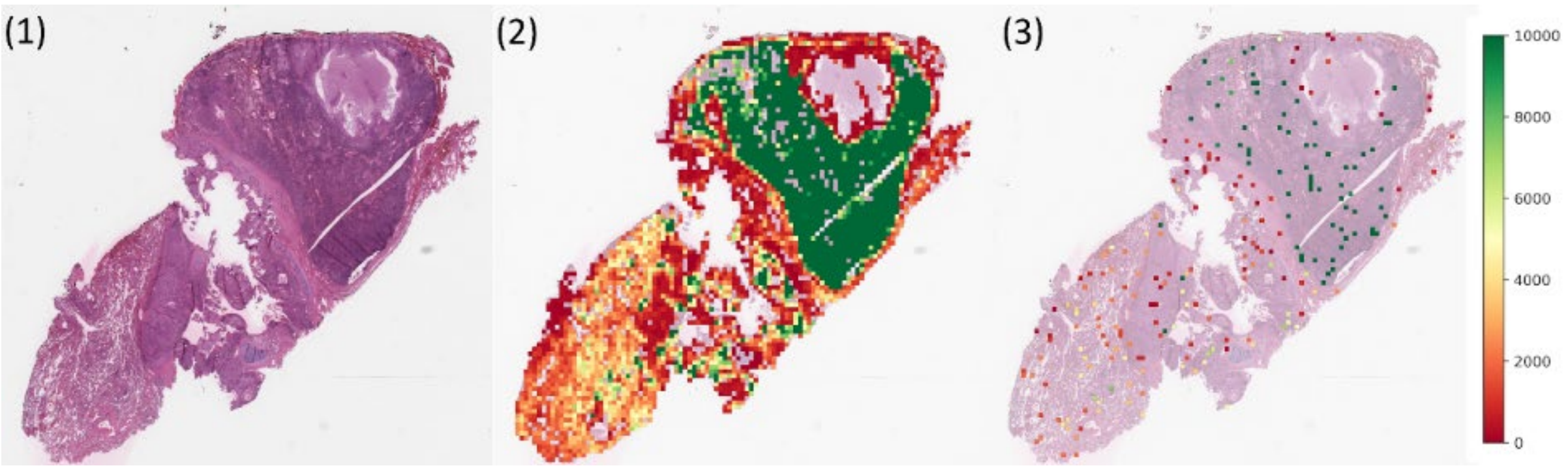

*Figure 2. Per patch TILs score visualization. (1) Original WSI, (2) all patch candidates, (3) 5% of patch candidates. The heatbar corresponds to TIL density scores for the patches in (2) and (3).*

## 2.5 Pipeline implementation setup

We implement our pipeline in Python v3.10 [23] and PyTorch v2.1.0 [24]. To efficiently access and manipulate data from WSIs, we use the OpenSlide v4.0.0 [25], libvips v8.15.0 [26], opencv v4.9.0 [27] and scikit-image v0.22.0 [28] libraries. The pipeline has a modular structure and can be parallelized at the WSI level, and even within steps. For instance, patch extraction can be distributed across a pool of processes, and the patch classification and cell quantification steps can utilize multiple GPUs to accelerate processing. However, for comparison and debugging purposes, parallel processing was not involved in validation runs. The development and validation were performed on a computer with an Intel Xeon W-2255 CPU with 10 cores, 128GB of RAM and an RTX Titan GPU with 24 GB VRAM. The source code of the pipeline and trained models are available on GitHub [29] and released under the MIT open source license.

## 2.6 Statistical Evaluation Between TILs, Clinical Variables and Prognosis

To investigate the prognostic potential of TILs in NSCLC, we perform statistical analyses to highlight relationships between TILs, clinicopathological characteristics and their combined impact on patient outcomes. The objective is to identify if our proposed TILs score can serve as a reliable indicator of patient survival and evaluate how it compares to the current standard IHC staining approach.

Given our pipeline's multi-stage nature, it is challenging to isolate and measure each stage's impact on patient survival separately. However, the pipeline's overall performance can be obtained by estimating the prognostic impact of the final TILs score. For ease of interpretation in survival analyses and for comparison with clinicopathological variables, we quantize the TILs score into four groups (Q1, Q2, Q3 and Q4) based on quantile cut-offs from the test subset of UNN-NSCLC WSI.

Disease-specific survival, defined as the time from diagnosis to lung cancer related death, is the chosen endpoint for survival analyses. Survival curves are calculated using the Kaplan Meier method and the difference between groups is tested using the log-rank test. The prognostic performance of the obtained TIL score, relative to cell counts for different immune cell subsets, is quantified using the concordance index (c-index), which serves a role analogous to the AUROC by measuring the model's discriminative ability in correctly ranking patients according to survival risk.

To assess the prognostic impact of our TILs score in relation to other clinical variables, we create multivariable models using the Cox proportional hazards models. The variables included in the models are pathological stage (pStage), differentiation grade, and TILs score, divided into quartiles. These variables were selected for inclusion based on their significance levels in univariate analyses using log-rank tests. In this analysis, the lowest category for each variable is used as the reference group. It allows for the hazard ratios (HRs) of the other categories to be interpreted relative to this baseline, facilitating meaningful comparisons of risk associated with each factor. Relations between TILs score, clinicopathological variables and dichotomized cell counts for different immune cell subsets are calculated using the $\chi^2$ or Fishers exact test whenever appropriate.

Throughout this study, we adopt an alpha level of 0.05 to determine statistical significance. For our analyses, we use Pandas v2.2.0 [30], lifelines v0.27.8 [31] Python libraries, and R version 4.3.1 [32] with the survival v3.5-7 [33] library.

# 3. Results

The following sections provide a comprehensive description of the performance, computational efficiency, and prognostic value of our pipeline. Briefly summarized we investigated patient level TILs scores in relation to prognosis to demonstrate its potential impact on lung cancer treatment.

## 3.1 Patch extraction

The patch extraction process is quantitatively assessed by two key metrics: the proportion of patches excluded, and the execution time required for the extraction procedure. Initially, the algorithm filters out approximately 70% of the total patches, which are identified as non-informative due to their absence of tissue or insufficient cellular content. This preliminary reduction is executed with an average processing time of approximately 5 minutes per WSI, corresponding to a pipeline performance of 20 patches per second being processed using a single-threaded process. Figure 3 illustrates a representative example of eligible patch extraction candidates with 768 x 768 size.

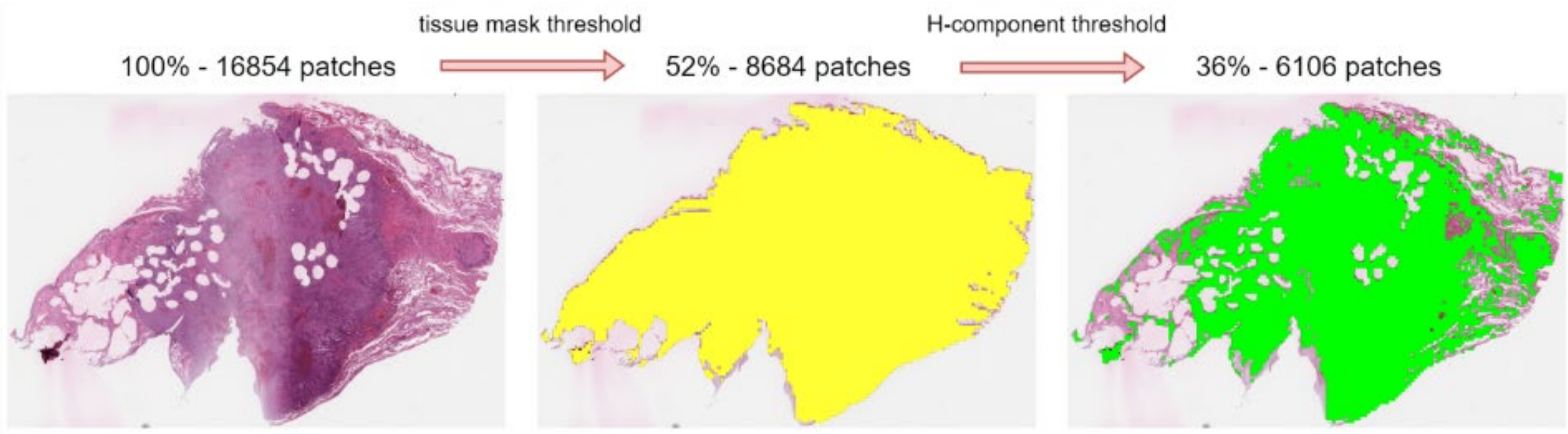

*Figure 3. Patch extraction reduction process with tissue mask and H-component thresholding*

### 3.2 Patch classification

The patch classification step aims to further reduce the subset of patches by excluding prognostically irrelevant regions. This step's performance is evaluated by its classification accuracy and impact on execution time for the subsequent pipeline stage.

We assess the accuracy of the patch classification process by comparing the predicted class labels with the ground truth labels in the test subset of the UNN-LC patch dataset. The test subset is clipped to the number of patches of the least represented class, leaving 295 patches of each class for testing. The model achieves an aggregated accuracy of 86.44% and multiclass Area Under the Receiver Operating Characteristic (average of the one-vs-rest AUCs) of 97.36%, demonstrating its effectiveness in distinguishing between patches with high density of necrotic cells or normal lung tissue and patches with tumor tissue and stroma. A comprehensive overview of the model's performance on the test subset is provided by the confusion matrix in Appendix Figure A2.1.

The patch classification process demonstrates its efficiency by quickly processing and classifying patches. On average, the patch classification model classifies 1000 patches in 60 seconds using the GPU, highlighting its potential for rapid analysis of lung cancer tissue.

The impact of the patch classification process on the pipeline's subsequent stages is significant. By assigning a class label to each patch and filtering out irrelevant patches, we reduce the number of quantified patches, while preserving the possibility to visually inspecting patches that are included and discarded for the subsequent cell quantification step. On average, 15% of the patches (necrotic and normal lung tissue) are discarded during this step. Figure 4 illustrates the patch classification step, showing examples of the patch classes found in the slide on the left (1) and eligible patches on the right (2).

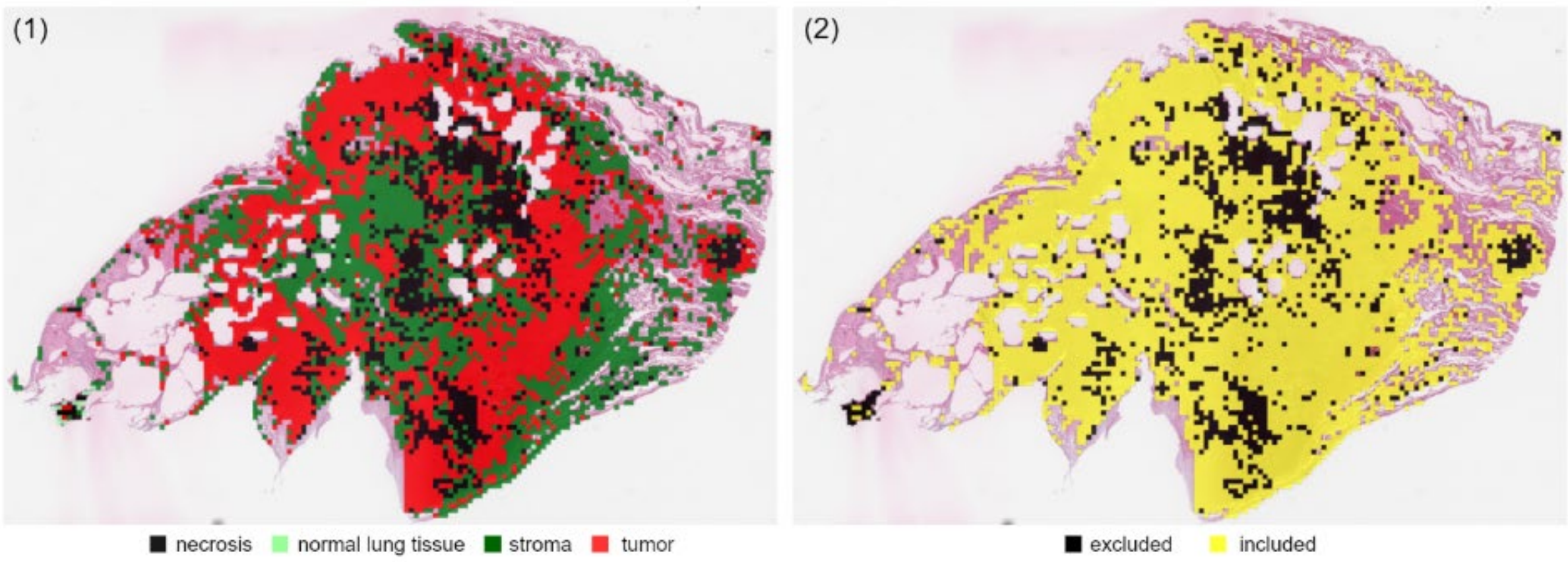

*Figure 4. Patch classification step visualization: (1) Color coded patch classification, (2) Patch selection, based on classes*

### 3.3 Cell quantification

The final stage of our pipeline is cell quantification, focusing on TILs. While the generation of cell contour masks for segmented cell instances is an optional feature, the process remains computationally intensive, due to the three separate decoding branches of the HoVer-Net architecture, along with the detailed post-processing of the resulting pixel maps. Our pipeline has been designed to automatically decrease the number of patches requiring analysis, so this reduction helps to mitigate the intensive computational demands of this stage.

We use a modified version of the HoVer-Net model, trained on the PanNuke dataset for this purpose. The original model's performance is well described in Shvetsov et al. 2022 [8]. We evaluate the model's performance using Dice score (DICE), achieving 82.48%, Jaccard score (or also known as intersection over union (IoU)) of 90.87%, and panoptic quality score (PQ) of 52.07% on the PanNuke test fold. These metrics, defined in Appendix A1.1, confirm the model's capability in delineating and classifying cellular structures.

By normalizing the absolute TIL counts and calculating the cell density for each patch, we can obtain a comprehensive TILs score for each WSI. This score can be used for further analysis and prognosis. In Figure 1 we can see three examples of 768x768 patches – one patch with just a few TILs, a patch with mean density of TILs and the patch with maximum TILs score from a single WSI. The score represents TILs density, given patch size, raw TILs counts and slide resolution.

### 3.4 Pipeline evaluation

To evaluate the clinical value of the pipeline, we use the test subset of the UNN-NSCLC dataset comprising 429 WSIs from patients that did not contribute patches used to train the patch classifier.

First, we compare the overall prognostic ability of the TILs score to that of the CD8 IHC score, using the c-index. The c-index helps us understand how well a certain factor can help predict outcomes in a survival model. In our case, we are looking at the TILs score and the CD8 IHC score. The c-index goes from 0.5 (random) to 1.0 (perfect outcome prediction).

To evaluate the randomized patch sampling, we use a subset of 50 WSIs, with assigned unique seed values to guarantee a diverse selection of patches, run the analyses and observe stable c-index values (0.65 ± 0.01) when 5% of patches are analyzed. Analyses of a higher volume of patches offer no additional prognostic benefit (Appendix Table A2.1). Hence, the gains in processing speed are not at the expense of prognostic information. Secondly, our observations reveal a linear relationship between processing times and patch ratio, presented in Table 1. Summarizing, we can obtain the same prognostic value, while being able to decrease processing times by 95%.

| **Analyzed Patch Ratio** | **Patch Extraction** | **Patch Classification** | **Patch Quantification** | **Total** |
|---|---|---|---|---|
| 100% | 5 minutes 23 sec | 4 minutes 44 sec | 72 minutes 32 sec | 82 minutes 39 sec |
| 70 % | 3 minutes 47 sec | 3 minutes 22 sec | 50 minutes 45 sec | 57 minutes 54 sec |
| 50 % | 2 minutes 42 sec | 2 minutes 24 sec | 36 minutes 15 sec | 41 minutes 21 sec |
| 20 % | 1 minute 5 sec | 57 seconds | 14 minutes 30 sec | 16 minutes 32 sec |
| 5% | 18 seconds | 15 seconds | 3 minutes 37 sec | 4 minutes 10 sec |

*Table 1. Average time performance of pipeline steps at different patch analysis ratios*

For the test subset of the UNN-NSCLC WSI dataset comprising 429 patients, the TILs score for 5% of patch candidates achieves a c-index of 0.649, while the CD8 IHC scoring method reaches a c-index of 0.599.

In the validation process using independent cohorts from TCGA, we observe variances in the pipeline performance. For TCGA-LUAD cohort with 387 slides, we obtain a c-index of 0.553, and for TCGA-LUSC with 382 slides, a c-index of 0.561.

### 3.5 Prognostic evaluation of TILs score

Table 2 presents the multivariable Cox proportional hazards analysis, evaluating the TILs score as a prognostic factor for disease-specific survival. In the test subset (n = 429), higher TILs scores indicate improved disease-specific survival. Specifically, patients in the second (Q2), third (Q3), and fourth (Q4) quartiles of the TILs score demonstrated progressively lower hazard ratios (HRs) compared to the reference group (Q1). The HRs were 0.60 (95% CI: 0.41–0.88, p = 0.008) for Q2, 0.50 (95% CI: 0.33–0.76, p = 0.001) for Q3, and 0.28 (95% CI: 0.17–0.45, p < 0.001) for Q4. This trend of decreasing hazard ratios with increasing TILs scores implies that patients with higher TILs scores had a significantly reduced risk of disease-specific death. Subgroup analyses by histology further reinforced these findings. In the lung adenocarcinoma (LUAD) subgroup (n = 189), higher TILs scores indicate better survival outcomes. The HRs for Q2, Q3, and Q4 were 0.44 (95% CI: 0.25–0.77, p = 0.004), 0.40 (95% CI: 0.22–0.71, p = 0.002), and 0.26 (95% CI: 0.13–0.53, p < 0.001), respectively, indicating a significant reduction in the risk of disease-specific death across all higher quartiles compared to Q1. In the LUSC subgroup we observe a trend similar to the LUAD subset.

| | Test subset (LUSC and LUAD) | | LUSC subset | | LUAD subset | |
|---|---|---|---|---|---|---|
| | HR (95% CI) | p | HR (95% CI) | p | HR (95% CI) | p |
| Differentiation | | | | | | |
| Reference - Poor | 1.0 | | 1.0 | | 1.0 | |
| Moderate | 0.88 (0.64-1.22) | 0.453 | 0.77 (0.48-1.23) | 0.271 | 1.05 (0.66-1.68) | 0.831 |
| Well | 0.41 (0.23-0.73) | 0.002 | 0.41 (0.14-1.19) | 0.103 | 0.34 (0.17-0.7) | 0.003 |
| pStage | | | | | | |
| Reference - I | 1.0 | | 1.0 | | 1.0 | |
| II | 1.55 (1.03-2.31) | 0.033 | 1.44 (0.77-2.68) | 0.249 | 1.6 (0.93-2.76) | 0.088 |
| III | 3.84 (2.62-5.64) | <0.001 | 4.59 (2.56-8.25) | <0.001 | 4.02 (2.32-6.96) | <0.001 |
| TILs score | | | | | | |
| Reference - Q1 | 1.0 | | 1.0 | | 1.0 | |
| Q2 | 0.6 (0.41-0.88) | 0.008 | 0.79 (0.44-1.41) | 0.419 | 0.44 (0.25-0.77) | 0.004 |
| Q3 | 0.5 (0.33-0.76) | 0.001 | 0.62 (0.32-1.2) | 0.159 | 0.4 (0.22-0.71) | 0.002 |
| Q4 | 0.28 (0.17-0.45) | <0.001 | 0.29 (0.14-0.6) | <0.001 | 0.26 (0.13-0.53) | <0.001 |

*Table 2. Multivariate Cox proportional hazards regression analysis of disease-specific survival*

To determine whether the addition of the TILs score improved the predictive performance of the multivariate Cox regression model, we perform a likelihood ratio test. As shown in Appendix Table A3.1, adding the TILs score significantly improved the model fit for the test subset ($\chi^2$ = 24.159, p < 0.001), the LUSC subset ($\chi^2$ = 12.006, p < 0.001), and the LUAD subset ($\chi^2$ = 7.2584, p = 0.007). Additional analysis of TILs score prognostic value is described in Appendix A3.1.

Kaplan-Meier survival analysis provides visual support for our findings, with a notable difference between the prognostic markers. The Kaplan-Meier curves for CD8 IHC score (Figure 5) display a significant degree of overlap among the various risk groups, which could imply a limitation in their prognostic utility. In contrast, the Kaplan-Meier curves for the TILs score (Figure 6) demonstrate a more pronounced separation between risk levels. This separation is especially prominent when comparing the highest and lowest TILs score quartiles, suggesting that the TILs score offers better prognostic value.

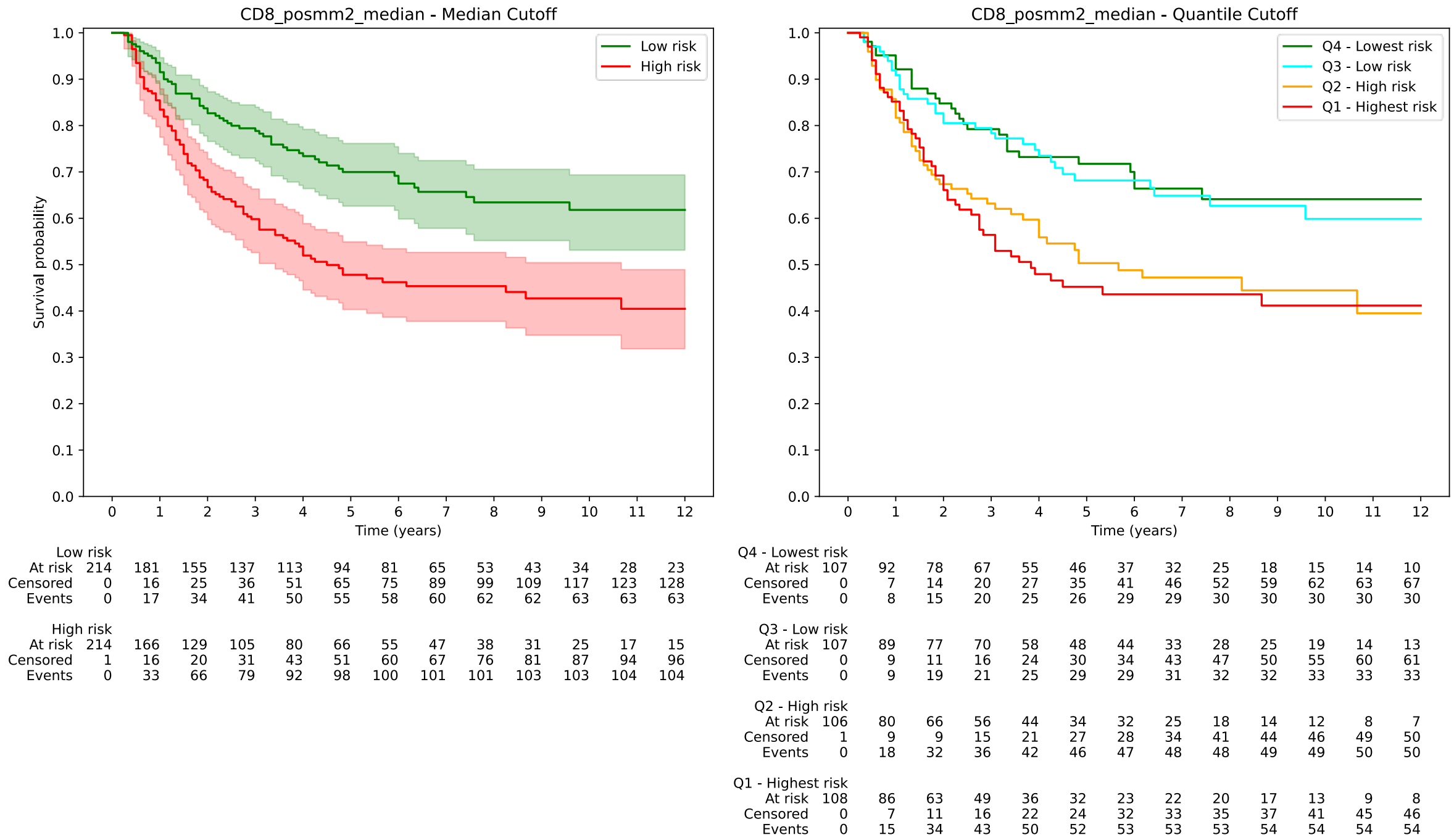

*Figure 5. Kaplan–Meier curves for CD8 IHC score*

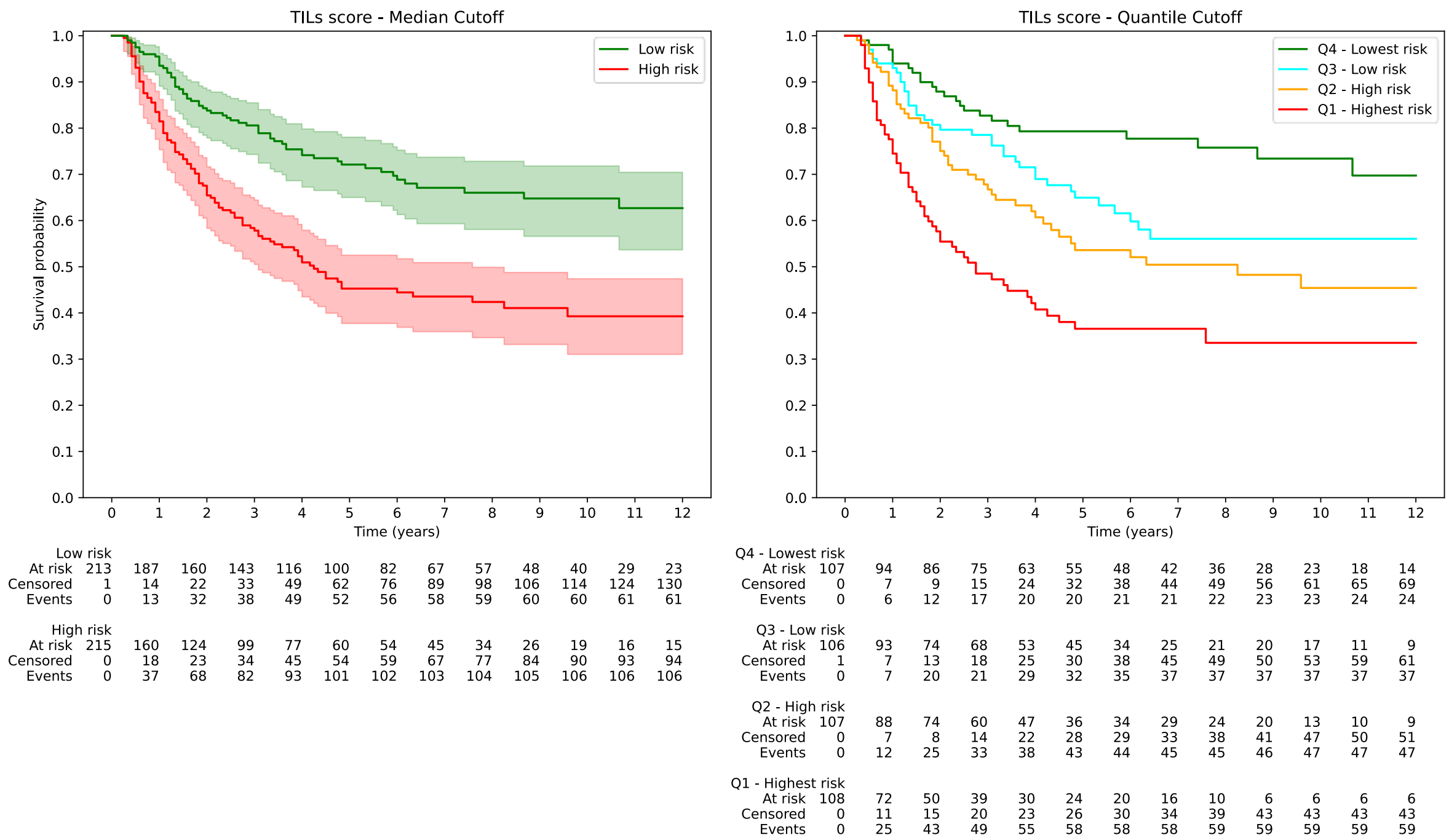

*Figure 6. Kaplan–Meier curves for TILs score*

# 4. Discussion

In this study, we have developed an automated, fast, and resource-efficient pipeline for analyzing high cell density patches with prognostic information from WSIs of lung cancer tissue. The pipeline integrates computer vision techniques for patch extraction with advanced deep learning models for classification and cell quantification, significantly contributing to computational pathology and NSCLC prognostication. Our pipeline performs favorably compared to other TIL evaluation approaches in NSCLC.

Our results demonstrate the pipeline's efficacy in selecting informative patches, quantifying cells, and approximating TIL density. The patch extraction step efficiently delineates tissue areas, identifying patches with sufficient cell counts. The classification accurately distinguishes between patches containing necrosis or normal lung tissue versus tumor tissue or tumor-associated stroma. Our cell quantification can delineate cellular structures, achieving high scores on internal validation data.

We also perform an ablation experiment to assess the impact of the patch classification step. Including a patch filtering process enhances the c-index by an average of 0.05, emphasizing its role in isolating high-value prognostic patches.

Determining the optimal ratio of patches for TIL evaluation is crucial. An optimal threshold at approximately 5%, averaging 350 patches per WSI, yields a c-index of approximately 0.65 with minimal standard deviation (Appendix Table A2.1). This threshold balances computational load and prognostic accuracy for our cohort data but warrants testing with other data sources. The computational time reduction observed at the 5% threshold is substantial when compared to the full analysis at the 100% patch ratio. Specifically, the average total time decreases from 82 minutes 39 seconds for the complete analysis to just 4 minutes 10 seconds. This represents an approximate 95% reduction in computational time, underscoring the efficiency gains achieved through the implementation of the proposed randomized sampling process. Although this observation may not hold for single WSI analysis due to variability in initial patch numbers, influenced by factors such as WSI dimensions, artifact presence, and necrotic area extent.

The modular pipeline design allows potential parallelization at both the WSI level and within individual steps, enhancing execution time. Additionally, explanations in the form of masks and overlays can assist pathologists in identifying suspicious cases, potentially improving prognostic accuracy.

Evaluation results indicate that the TILs score is a significant prognostic factor and can enhances the predictive ability of models that include just pStage and differentiation grade, which are well-established prognostic factors for lung cancer. We also demonstrate that the TILs score is a strong independent prognostic factor in non-small cell lung cancer, particularly in LUAD patients (Table 2).

The significant inverse relationship between TILs scores and the risk of disease-specific death highlights the critical role of the immune response in cancer progression. Patients with higher TILs scores, reflecting greater infiltration of immune cells within the tumor, tend to have more favorable outcomes. This pattern suggests that quantifying TILs can effectively stratify patients based on their immune response to the tumor, providing valuable prognostic information that could inform personalized treatment strategies.

Despite promising results, our study has limitations. The multi-stage pipeline poses challenges in isolating and measuring the individual stage's impacts on the final result. Additionally, some low-information patches are still included due to H-component thresholding instability in certain scenarios like tissue folding and pen marker artifacts [34]. Moreover, the generalization capabilities of convolutional neural networks in a limited data environment remains an issue. The models may not capture nuanced features in data-scarce settings, leading to potential overfitting and reduced predictive performance on unseen datasets. Furthermore, misclassification of necrotic or normal lung regions can lead to underestimations, influencing TIL density scores. The confusion matrix (Appendix Figure A2.1) shows a higher misclassification rate for the 'necrosis' class.

While the results from both TCGA-LUAD and TCGA-LUSC datasets fall below expectations, this disparity underscores the challenges associated with datasets that exhibit significant variability in quality and preparation protocols. This highlights the difficulty of achieving reliable performance due to variations in data quality, annotation standards, imaging protocols, scanner configurations, and tissue sample preparation methods. However, it is plausible that by fine-tuning the pipeline for a specific clinical setting, where such variance is likely to be reduced, these issues can be mitigated. Applying the pipeline to the more uniform conditions of a particular clinic may yield improved and more reliable performance, suggesting that the observed challenges may be less impactful in a practical context.

More complex deep learning models, such as those trained in a self-supervised manner, could reduce reliance on labels and thus enhance generalizability [35], [36]. Using foundational models trained on large datasets may offer additional improvements [37], [38], [39]. While these experiments are beyond this paper's scope, they pose reasonable directions for the future research.

Future work could also refine patch sampling to reduce irrelevant patches and focus on regions such as tumor borders, where TILs may indicate a stronger immune response. Including other cell types with prognostic properties in NSCLC is also recommended. A detailed computational cost analysis, including energy consumption, which is an important aspect of using deep learning approaches [40], [41], and performance comparisons across diverse hardware configurations, would provide a holistic view of our approach's cost-effectiveness and scalability in different clinical settings.

# 5. Conclusions

In conclusion, this study introduces a novel, automated pipeline for efficient TILs evaluation in WSIs. Our results suggest that the TILs score, generated by our pipeline, offers improved prognostic performance compared to the CD8 IHC score. This is evidenced by the superior c-index and the more distinct separation of high and low risk groups in the Kaplan-Meier survival curves generated by our TILs score. The pipeline also provides explainable heatmaps and patch visualizations across WSIs.

The developed pipeline integrates computer vision techniques and state-of-the-art deep learning models for efficient patch extraction, classification, and cell quantification. Specifically, the pipeline maintains high accuracy, while also demonstrating computational efficiency and speed. This is achieved despite the large volume of patches contained in the WSIs and the complex models required to accurately quantify overlapping cells in these images. The ability to process and evaluate patches on-the-fly, without the need to save them to disk, coupled with reduced number of patches for cell quantification, underscores the pipeline's potential value in performance when compared to current clinical approach in terms of speed and computational resource utilization.

Importantly, the TILs score can serve as a better prognostic marker, enabling faster and more informed decisions by pathologists. This facilitates the development of more accurate and personalized treatment strategies for lung cancer patients, potentially improving current clinical practices.

Further studies are needed to validate these findings. The clinical applicability of our pipeline also needs to be assessed in more detail. Future research should focus on these aspects, exploring the full potential of our approach by investigating other biomarkers for improving NSCLC prognosis and transforming the landscape of computational pathology.

## Acknowledgments

This work was funded in part by the Research Council of Norway grant no. 309439 SFI Visual Intelligence, and the North Norwegian Health Authority grant no. HNF1521-20.

# Appendix

## A1. Supplementary Methods

### A1.1 Definition of metrics used to evaluate the performance of the cell segmentation model

The results of our cell segmentation pipeline are detailed in Shvetsov et al 2022. Briefly, we used the following definitions of metrics to evaluate the cell segmentation pipeline:

1) $DICE = \frac{2*TP}{2*TP+FP+FN}$ , where TP and FP represent the number of true positive and false positive pixels, respectively.
2) $\mathrm{IoU} = \frac{\sum_{i=1}^{n}|X_i \cap Y_i|}{\sum_{i=1}^{n}|X_i \cup Y_i|}$ , where n represents the number of classes, where $X_i$ represents the set of pixels predicted to belong to class i, and $Y_i$ represents the set of pixels that belong to class i in the ground truth.
3) $PQ = \frac{IOU}{TP+0.5FP+0.5FN}$ , where TP, FP and FN represent the number of true positives, false positive and false negative pixels, respectively.

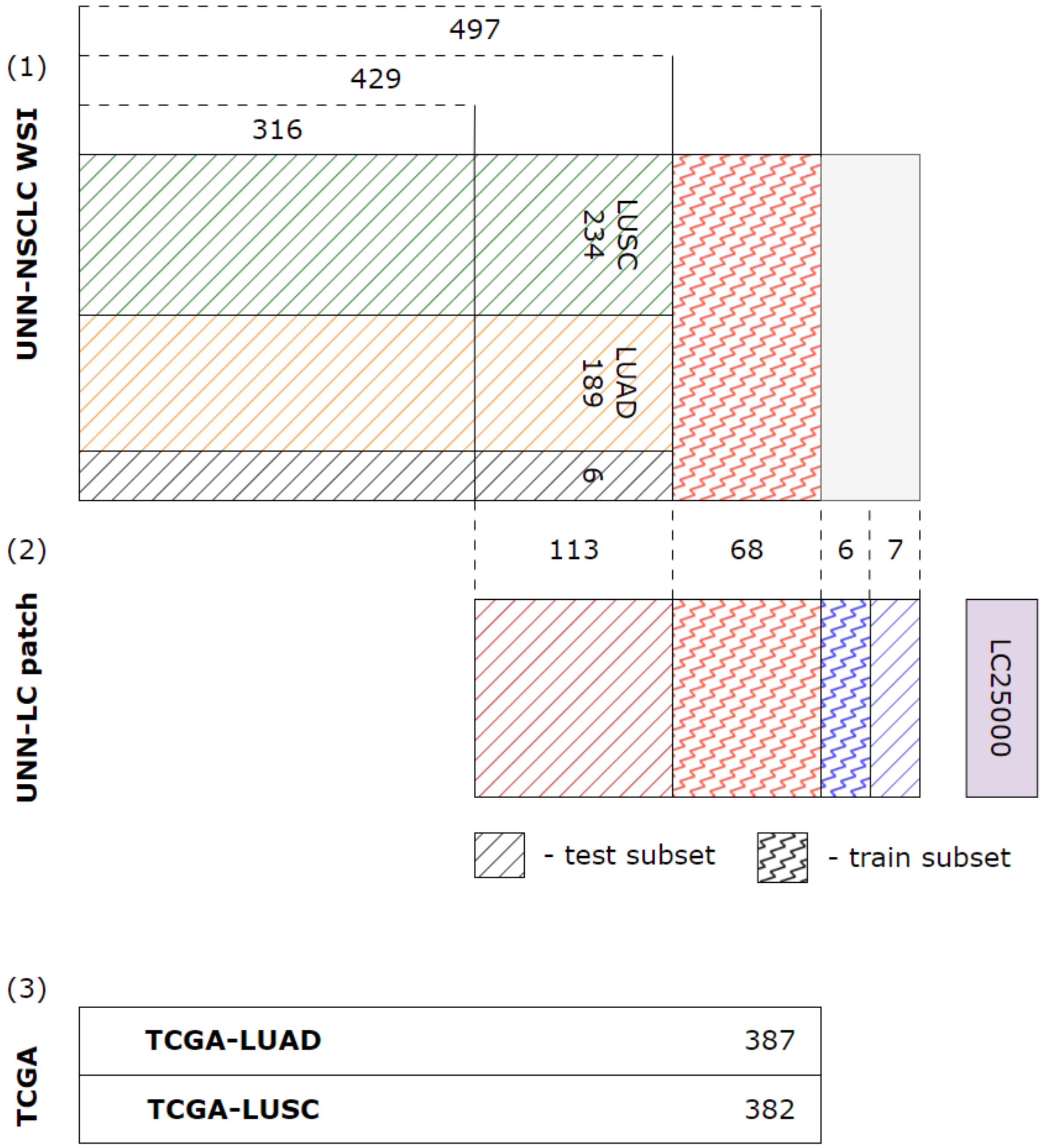

*Figure A1.1 Distribution and relations of WSIs used in training and testing of the pipeline for (1) UNN-NSCLC WSI dataset, (2) UNN-LC patch dataset and (3) TCGA dataset*

## A2. Supplementary Results

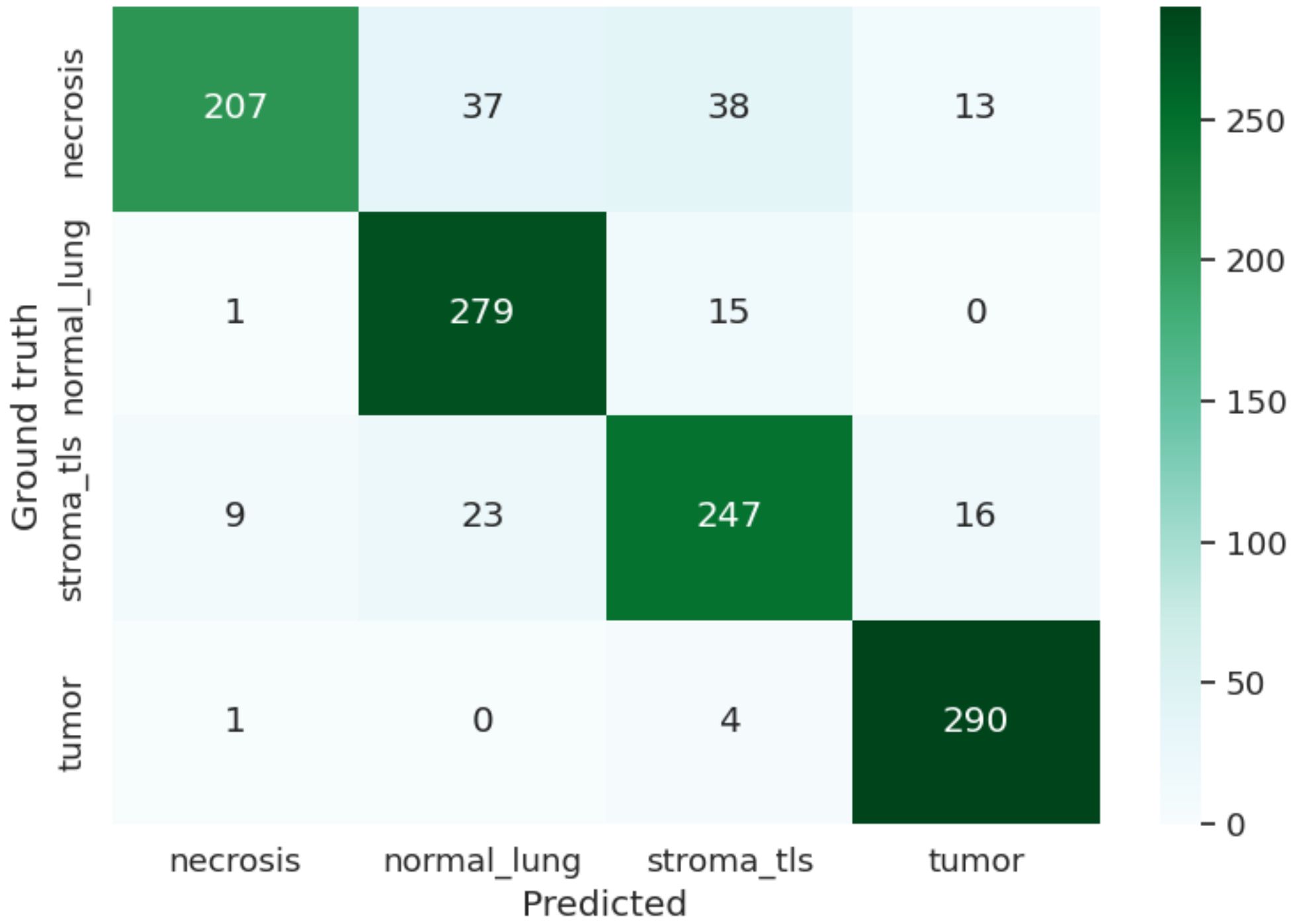

*Figure A2.1 Confusion matrix of the classification model for the test subset*

| Patch ratio | c-index (average) | Standard deviation | Number of patches (average) |
|---|---|---|---|
| 0.005 | 0.63214 | 0.0113 | 21 |
| 0.01 | 0.63904 | 0.0083 | 42 |
| 0.05 | 0.64851 | 0.0035 | 215 |
| 0.1 | 0.65090 | 0.0023 | 432 |
| 0.2 | 0.65144 | 0.0017 | 864 |
| 0.3 | 0.65177 | 0.0014 | 1297 |
| 0.4 | 0.65209 | 0.0010 | 1730 |
| 0.5 | 0.65190 | 0.0009 | 2163 |

*Table A2.1 Analyzed patch ratios and corresponding c-index calculated using Monte-Carlo simulation and 100 iterations*

## A3. Supplementary Validation

### A3.1 Additional TILs score analysis

In our additional analyses, we examine the relationships between the TILs score and clinicopathological variables to assess potential confounding factors and validate the independence of the TILs score as a prognostic marker (Appendix Table A3.2). We use chi-squared tests when the expected counts were sufficiently large ($E_c >= 5$) and applied Fisher's exact tests in other cases, to identify any significant associations. In the test subset (n = 429), our analyses revealed no significant associations between the TILs score and key clinicopathological variables such as age, gender, weight loss, smoking status, ECOG performance status, pStage, differentiation, and vascular invasion ($p > 0.05$). The high p-values suggest that the TILs score is independent of these potential confounders and can potentially be unbiased prognostic factor.

Further exploring the biological value of the TILs score, we examined its relationship with the expression levels of IHC-based markers of common immune cells: CD3 (pan T-cells), CD4 (helper T-cells), CD8 (cytotoxic T-cells), and CD20 (B-cells), as presented in Appendix Table A3.2. Patients in the highest TILs score quartile (Q4) had a significantly greater proportion of high CD8 expression compared to those in the lowest quartile (Q1). Cochran-Armitage tests for trend were conducted to assess the presence of trends in TILs score quartiles and the expression levels of the immune cell markers. The test results demonstrated significant positive trends for all markers, with the most pronounced observed for CD3 ($Z = 7.16$, $p < 0.001$) and the least for CD20 ($Z = 3.43$, $p < 0.001$). These findings suggest that the TILs score, as a measure of immune cell infiltration, exhibits a monotonously growing trend across clinical markers from Q1 to Q4.

To evaluate the prognostic value of the TILs score in a univariate model, we examined both the TILs score and the CD8 IHC score separately for predicting disease-specific survival in NSCLC patients, as detailed in Appendix Table A3.3. In the test subset, patients in the highest TILs score quartile (Q4) had a significantly reduced hazard of disease-specific death compared to those in the lowest quartile (Q1), with HR of 0.26 (95% CI: 0.16–0.40). The CD8 IHC score also showed a trend towards improved survival in higher quartiles, though the hazard was higher for Q4 with HR of 0.46 (95% CI: 0.30–0.70).

We also perform Kaplan-Meier survival analyses stratified on histology (Appendix Figure A3.1) and pStage (Appendix Figure A3.2). In histologically differentiated patients, Kaplan-Meier survival analyses demonstrate a distinct stratification between the lowest and highest quartiles of TILs scores in both LUAD and LUSC. For pStage, we observe a consistent trend where higher TILs scores are linked to better survival outcomes across all stages. The most pronounced separation between risk groups is observed in early-stage tumors.

| | Test subset (LUSC and LUAD) | | LUSC subset | | LUAD subset | |
|---|---|---|---|---|---|---|
| | Baseline | Extended | Baseline | Extended | Baseline | Extended |
| likelihood ratio | -903.80 | -891.72 | -358.31 | -352.31 | -411.72 | -408.09 |
| $\chi^2$ | 24.159 | | 12.006 | | 7.2584 | |
| p | $8.87 \times 10^{-7}$ | | 0.0005304 | | 0.007057 | |

*Table A3.1 Likelihood Ratio Test results comparing Baseline (pStage + Differentiation) and Extended (pStage + Differentiation + TILs) cox regression models*

| | All patients (n=497) | | | | | Test patients (n=429) | | | | | Train patients (n=68) | | | | |
|---|---|---|---|---|---|---|---|---|---|---|---|---|---|---|---|
| | TILs score | | | | | TILs score | | | | | TILs score | | | | |
| | Q1 | Q2 | Q3 | Q4 | p | Q1 | Q2 | Q3 | Q4 | p | Q1 | Q2 | Q3 | Q4 | p |
| Age | | | | | | | | | | | | | | | |
| <65 | 50 | 57 | 47 | 53 | 0.649 | 47 | 45 | 41 | 45 | 0.880 | 3 | 12 | 6 | 8 | 0.393 |
| ≥65 | 62 | 76 | 80 | 72 | | 61 | 62 | 66 | 62 | | 1 | 14 | 14 | 10 | |
| Gender | | | | | | | | | | | | | | | |
| Female | 31 | 41 | 43 | 45 | 0.542 | 30 | 37 | 36 | 41 | 0.431 | 1 | 4 | 7 | 4 | 0.441 |
| Male | 81 | 92 | 84 | 80 | | 78 | 70 | 71 | 66 | | 3 | 22 | 13 | 14 | |
| Weight loss | | | | | | | | | | | | | | | |
| <10% | 103 | 11 | 110 | 117 | 0.268 | 99 | 97 | 90 | 101 | 0.118 | 4 | 20 | 20 | 16 | 0.094 |
| ≥10% | 9 | 16 | 16 | 8 | | 9 | 10 | 16 | 6 | | 0 | 6 | 0 | 2 | |
| Smoking status | | | | | | | | | | | | | | | |
| Never smoked | 5 | 4 | 1 | 7 | 0.113 | 5 | 2 | 0 | 6 | 0.106 | 0 | 2 | 1 | 1 | 0.294 |
| Present smoker | 64 | 80 | 89 | 83 | | 63 | 66 | 75 | 69 | | 1 | 14 | 14 | 14 | |
| Previous smoker | 43 | 49 | 37 | 35 | | 40 | 39 | 32 | 32 | | 3 | 10 | 5 | 3 | |
| ECOG status | | | | | | | | | | | | | | | |
| Normal | 57 | 76 | 77 | 84 | 0.110 | 55 | 66 | 65 | 70 | 0.284 | 2 | 10 | 12 | 14 | 0.124 |
| Slightly reduced | 42 | 47 | 43 | 37 | | 41 | 35 | 36 | 33 | | 1 | 12 | 7 | 4 | |
| In bed <50% | 13 | 10 | 7 | 4 | | 12 | 6 | 6 | 4 | | 1 | 4 | 1 | 0 | |
| Histology | | | | | | | | | | | | | | | |
| LUSC | 62 | 70 | 63 | 82 | 0.085 | 59 | 56 | 51 | 68 | 0.129 | 3 | 14 | 12 | 14 | 0.619 |
| LUAD | 50 | 61 | 61 | 41 | | 49 | 50 | 53 | 37 | | 1 | 11 | 8 | 4 | |
| Other | 0 | 2 | 3 | 2 | | 0 | 1 | 3 | 2 | | 0 | 1 | 0 | 0 | |
| pStage | | | | | | | | | | | | | | | |
| IA | 26 | 33 | 33 | 41 | 0.117 | 25 | 27 | 30 | 35 | 0.157 | 1 | 6 | 3 | 6 | 0.875 |
| IB | 17 | 20 | 29 | 12 | | 16 | 15 | 24 | 11 | | 1 | 5 | 5 | 1 | |
| IIA | 8 | 11 | 14 | 12 | | 7 | 8 | 11 | 10 | | 1 | 3 | 3 | 2 | |
| IIB | 25 | 32 | 25 | 37 | | 25 | 24 | 19 | 32 | | 0 | 8 | 6 | 5 | |
| IIIA | 32 | 31 | 20 | 20 | | 31 | 27 | 17 | 16 | | 1 | 4 | 3 | 4 | |
| IIIB | 4 | 6 | 6 | 3 | | 4 | 6 | 6 | 3 | | 0 | 0 | 0 | 0 | |
| Differentiation | | | | | | | | | | | | | | | |
| Poor | 47 | 51 | 58 | 50 | 0.693 | 46 | 42 | 49 | 42 | 0.872 | 1 | 9 | 9 | 8 | 0.635 |
| Moderate | 52 | 61 | 53 | 52 | | 49 | 48 | 43 | 46 | | 3 | 13 | 10 | 6 | |
| Well | 13 | 21 | 16 | 23 | | 13 | 17 | 15 | 19 | | 0 | 4 | 1 | 4 | |
| Vascular invasion | | | | | | | | | | | | | | | |
| No | 91 | 113 | 102 | 104 | 0.725 | 88 | 90 | 88 | 88 | 0.930 | 3 | 23 | 14 | 16 | 0.322 |
| Yes | 21 | 19 | 24 | 20 | | 20 | 16 | 18 | 18 | | 1 | 3 | 6 | 2 | |

*Table A3.2 Quantized TILs score distribution with clinicopathological variables for UNN-NSCLC WSI dataset. Patient numbers depend on availability of clinical data from the respective subset.*

| | All patients (n=497) | | | | | | Test patients (n=429) | | | | | | Train patients (n=68) | | | | | |
|---|---|---|---|---|---|---|---|---|---|---|---|---|---|---|---|---|---|---|
| | TILs score | | | | | | TILs score | | | | | | TILs score | | | | | |
| | Q1 | Q2 | Q3 | Q4 | z | p | Q1 | Q2 | Q3 | Q4 | z | p | Q1 | Q2 | Q3 | Q4 | z | p |
| CD3 | | | | | | | | | | | | | | | | | | |
| [0,1e+03] | 103 | 109 | 89 | 60 | 7.68 | <0.001 | 99 | 90 | 73 | 54 | 7.16 | <0.001 | 4 | 19 | 16 | 6 | 2.76 | <0.05 |
| (1e+03,5e+03] | 9 | 23 | 37 | 61 | | | 9 | 17 | 33 | 51 | | | 0 | 6 | 4 | 10 | | |
| CD4 | | | | | | | | | | | | | | | | | | |
| [0,550] | 90 | 86 | 65 | 57 | 5.98 | <0.001 | 87 | 72 | 54 | 53 | 5.34 | <0.001 | 3 | 14 | 11 | 4 | 2.48 | <0.05 |
| (550,5e+03] | 21 | 44 | 60 | 68 | | | 20 | 34 | 51 | 54 | | | 1 | 10 | 9 | 14 | | |
| CD8 | | | | | | | | | | | | | | | | | | |
| [0,500] | 76 | 64 | 56 | 36 | 5.88 | <0.001 | 73 | 53 | 44 | 31 | 5.77 | <0.001 | 3 | 11 | 12 | 5 | 1.26 | 0.21 |
| (500,5e+03] | 36 | 69 | 71 | 89 | | | 35 | 54 | 63 | 76 | | | 1 | 15 | 8 | 13 | | |
| CD20 | | | | | | | | | | | | | | | | | | |
| [0,400] | 101 | 109 | 100 | 92 | 3.46 | <0.001 | 97 | 88 | 84 | 78 | 3.42 | <0.001 | 4 | 21 | 16 | 14 | 0.70 | 0.48 |
| (400,5e+03] | 10 | 23 | 27 | 33 | | | 10 | 18 | 23 | 29 | | | 0 | 5 | 4 | 4 | | |

*Table A3.3 Quantized TILs score distribution over IHC based immune cell marker subsets and Cochran-Armitage test results for trend (z).*

| | All patients (n=497) | | | | | Test patients (n=429) | | | | | Train patients (n=68) | | | | |
|---|---|---|---|---|---|---|---|---|---|---|---|---|---|---|---|
| | N (%) | 5 Year | Median | HR (95%CI) | p | N (%) | 5 Year | Median | HR (95%CI) | p | N (%) | 5 Year | Median | HR (95%CI) | p |
| CD8_posmm2_median | | | | | | | | | | | | | | | |
| Q1 | 124 (25) | 45 | 42 | 1 | <0.001 | 108 (25) | 45 | 45 | 1 | <0.001 | 16 (24) | 40 | 25 | 1 | 0.08 |
| Q2 | 123 (25) | 52 | 73 | 0.82 (0.54-1.25) | | 107 (25) | 50 | 68 | 0.88 (0.57-1.38) | | 16 (24) | 61 | NR | 0.47 (0.13-1.66) | |
| Q3 | 126 (25) | 71 | 235 | 0.45 (0.3-0.68) | | 107 (25) | 68 | NR | 0.49 (0.32-0.76) | | 19 (28) | 88 | 235 | 0.24 (0.07-0.82) | |
| Q4 | 124 (25) | 70 | NR | 0.5 (0.33-0.75) | | 107 (25) | 72 | NR | 0.46 (0.3-0.7) | | 17 (25) | 62 | 84 | 0.73 (0.2-2.58) | |
| TILs score | | | | | | | | | | | | | | | |
| Q1 | 112 (23) | 36 | 32 | 1 | <0.001 | 108 (25) | 37 | 32 | 1 | <0.001 | 4 (6) | 33 | 30 | 1 | 0.12 |
| Q2 | 133 (27) | 54 | 75 | 0.62 (0.4-0.96) | | 107 (25) | 54 | 98 | 0.61 (0.38-0.96) | | 26 (38) | 53 | 61 | 0.75 (0.09-6.41) | |
| Q3 | 127 (26) | 67 | NR | 0.41 (0.27-0.64) | | 107 (25) | 65 | NR | 0.45 (0.28-0.71) | | 20 (29) | 78 | NR | 0.28 (0.03-2.39) | |
| Q4 | 125 (25) | 78 | 235 | 0.27 (0.18-0.42) | | 107 (25) | 79 | NR | 0.26 (0.16-0.4) | | 18 (26) | 72 | 235 | 0.35 (0.04-2.96) | |

*Table A3.4 Quantized CD8 TILs density and TILs score as predictors of disease-specific survival in non-small cell lung cancer patients. HR and p-values were calculated based on 5-year survival. Abbreviations: NR, not reached.*

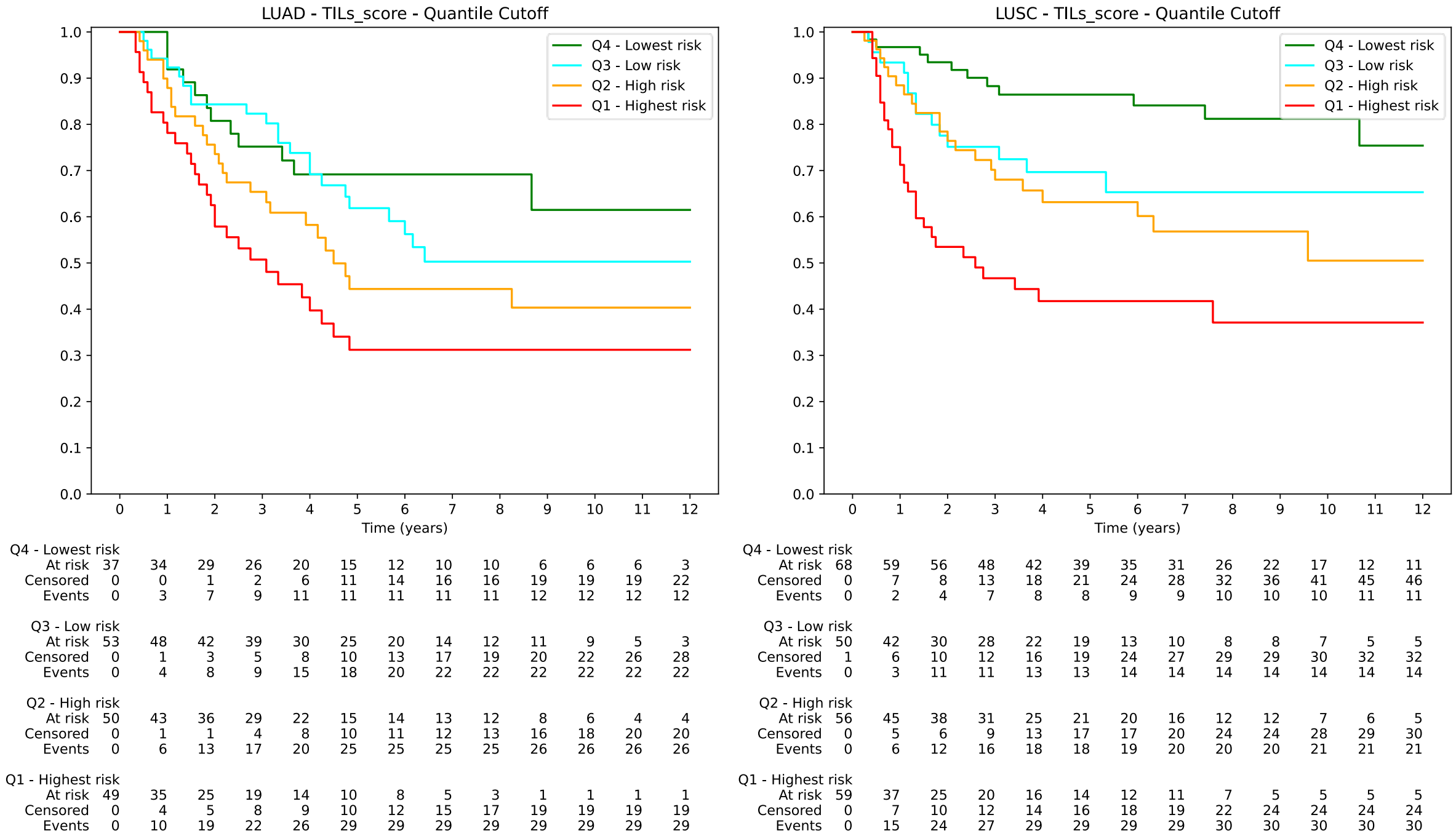

*Figure A3.1 KM curves for TILs score stratified by histology.*

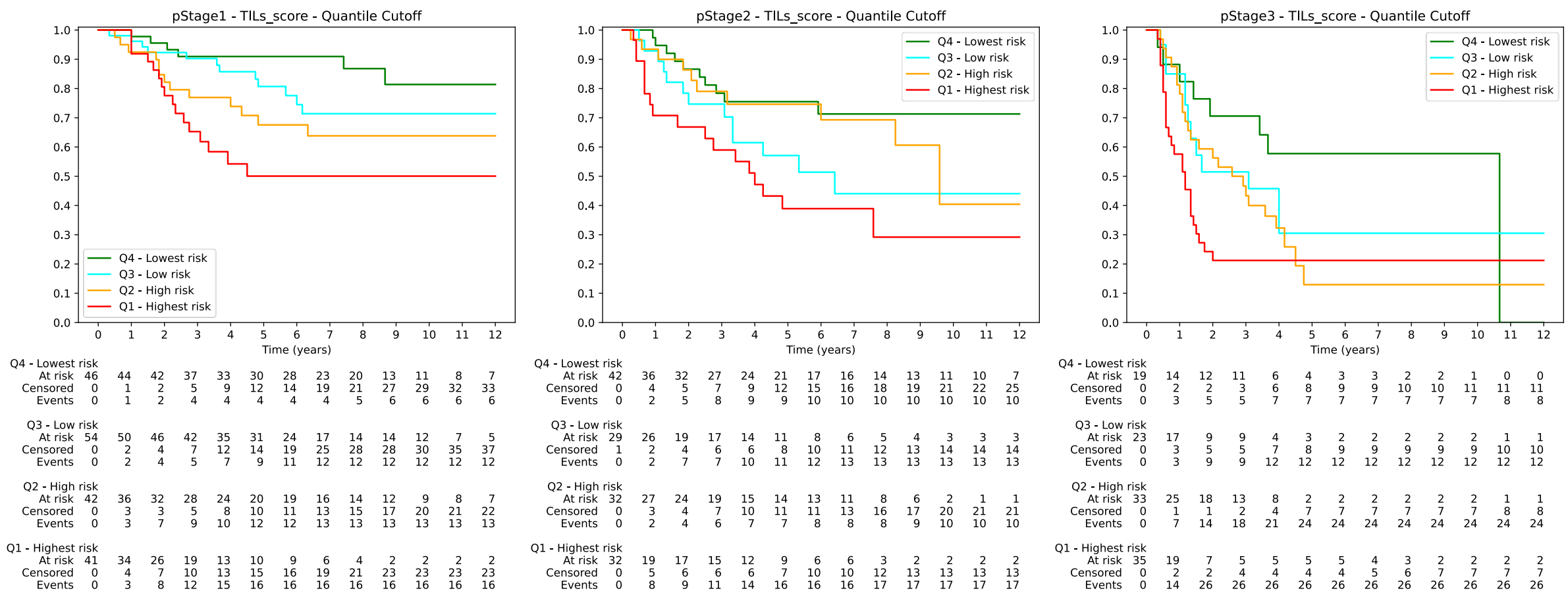

*Figure A3.2 KM curves for TILs score stratified by pStage*